# FAST AND ACCURATE, CONVOLUTIONAL NEURAL NETWORK BASED APPROACH FOR OBJECT DETECTION FROM UAV

Xiaoliang Wang
Department of Technology
College of Engineering and Technology
Virginia State University
Petersburg, VA, USA
xwang@vsu.edu

Peng Cheng
Department of Technology
College of Engineering and Technology
Virginia State University
Petersburg, VA, USA
pcheng@vsu.edu

Xinchuan Liu
Department of Technology
College of Engineering and Technology
Virginia State University
Petersburg, VA, USA
xliu@vsu.edu

Benedict Uzochukwu
Department of Technology
College of Engineering and Technology
Virginia State University
Petersburg, VA, USA
buzochukwu@vsu.edu

*Abstract*—Unmanned Aerial Vehicles (UAVs), have intrigued different people from all walks of life, because of their pervasive computing capabilities. UAV equipped with vision techniques, could be leveraged to establish navigation autonomous control for UAV itself. Also, object detection from UAV could be used to broaden the utilization of drone to provide ubiquitous surveillance and monitoring services towards military operation, urban administration and agriculture management. As the data-driven technologies evolved, machine learning algorithm, especially the deep learning approach has been intensively utilized to solve different traditional computer vision research problems. Modern Convolutional Neural Networks based object detectors could be divided into two major categories: one-stage object detector and two-stage object detector. In this study, we utilize some representative CNN based object detectors to execute the computer vision task over Stanford Drone Dataset (SDD). State-of-the-art performance has been achieved in utilizing focal loss dense detector RetinaNet based approach for object detection from UAV in a fast and accurate manner.

## I. INTRODUCTION

Unmanned aerial vehicles (UAVs), because of their capabilities of offering ubiquitous computing resources, have been adopted to provide services in different application scenario. Camera based vision systems could be configured over UAV to perform image processing and video streaming. Vision assisted UAV system could be leveraged to deploy navigation autonomous control for itself. For example, relying on the camera involved sensory systems, vision assisted UAV could achieve pathfinding, balance control and landing maneuvering in a adaptively controlled manner [1][2][3]. Also vision assisted UAV plays critical roles in military operations, e.g., military surveillance [4][5]. UAV equipped with computer vision techniques could be used to provide effective urban administrations. Chen et al. [6] in their work implement an UAV based Fog Computing system to leverage video processing for urban surveillance. UAV could also rely on computer vision techniques to deploy agriculture management [7][8][9]. Within many different application scenarios of the UAVs, the methodology of object detection from UAV plays critical roles, which is required to be deployed with high accuracy and efficiency.

Previously, fixed feature extraction methodologies have been employed to serve object detection from UAV [10][11]. As its the continuing evolution recently, Convolutional Neural Network (CNN) becomes another group of popular methods, which are able to provide UAV with the ability to detect and recognize objects from long distance [12][13][14]. Although CNN based methods are able to achieve relatively high accuracy compared to those traditional computer vision methods, they are sometimes not so satisfied considering the aspect of computation speed.

SSD [15], Faster R-CNN [16], Focal Loss Dense Detector RetinaNet [17] are chosen as representative CNN object detectors in this paper to undertake the computer vision task over Stanford Drone Dataset (SDD) [18]. State-of-the-art experiment result achieved by RetinaNet indicates that CNN based object detector could be well utilized by UAV to perform object detection efficiently and precisely.

## II. BACKGROUND

### A. Application of Vision Assisted UAV

Vision assisted UAV could achieve navigation and autonomous control for itself. Saripalli et al. [19] in their study explored the vision based strategies for autonomous control during UAV landing period. Scaramuzza et al. [20] incorporated vision techniques into their UAV design to enhance the navigation performance. PIXHAWK [21] is a popular flight control system designed with computer vision algorithms to execute the obstacle detection during UAV operation. Kendoul et al. [22] leveraged optical flow based

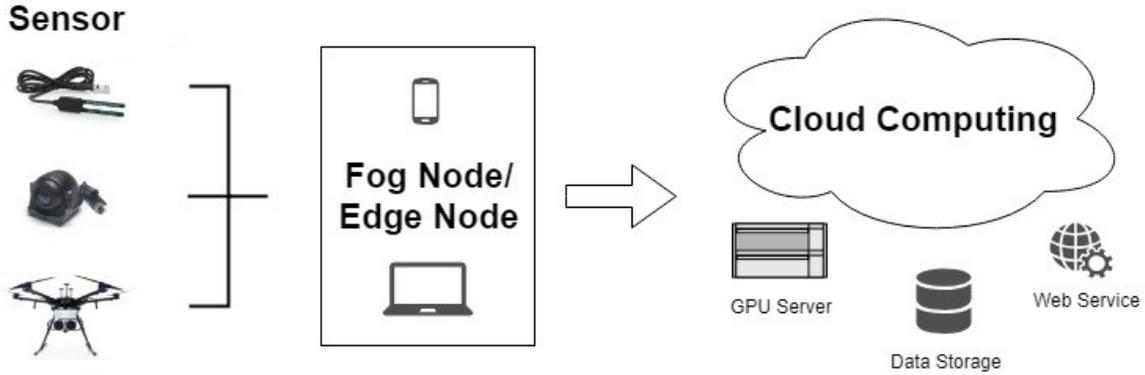

Figure 1. UAV Based Fog Computing Platform

computer vision techniques to implement self-maneuvering operations for aerial vehicle.

Vision assisted UAV could also offer video or image surveillance services in different application scenarios. Ryan et al. [23] in their work gave a general overview towards the employment of object detection involved computer vision techniques for aerial surveillance. Schneiderman [24] discussed the drone usage in military services. Avola et al. [25] deployed computer vision techniques from UAV to monitor the ever changing environment. In the work of Horton et al. [26], they make use of UAV for peach flower monitoring. Vanegas et al. [27] proposed methods to improve the effectiveness of plant pest surveillance. Albani et al. [28] developed swarm drone systems for plant (such as sugar beets and volunteer potatoes) monitoring. Lately, UAVs have also been dedicated to the service of smart cities for urban administration [29] [30].

*B. Object Detection Method*

Some traditional computer vision technologies, such as those handcraft feature extraction methods have been used for object detection [31] [32] [33]. There is also a series of study, which utilize handcraft features for object detection from UAV [5][34]. Until recently, deep learning has been proved to be able to achieve sophisticated result in undertaking different computer vision tasks [35]. Specifically, for the object detection task, convolutional neural network based object detectors could be divided into two categories: two-stage object detector and one-stage object detector. Compared to the one-stage object detector, two-stage object detector normally utilizes its first stage to generate the collection of Regions of Interest (RoI) proposals. Representatives of the two-stage object detector include R-CNN [36] and later evolve to Fast R-CNN [37] and Faster R-CNN [16]. For the one-stage object detector, some of the representative study includes SSD [15] and YOLO [38]. From the perspective of timing performance, one-stage detector present some faster speed result compared to two-stage detector. But one-stage detector is reported to have some sampling class imbalance issues. A large quantity of easy negatives brought about by the one-stage object detector contain much less information compared to the scarce positive samples; on the other side, those negative samples would also downgrade the efficiency of training procedures for deep learning. Focal loss [17] is introduced to solve this class imbalance issue.

*C. UAV Based System Architecture*

UAV could play different roles in a cyber physical system. Depending on the characteristics of the computation workload and wireless communication environment, the UAV could selectively only collect the sensory information and transmit all the data back to the ground for further processing. Rather than that, another option is allocate merely inference procedures of deep learning algorithms executed over UAV while the computation intensive training procedures would still be offloaded to the ground server. In that case, if deep learning involved procedures is decided to be executed over the resource limited devices, such as UAV, some further investigation towards the deep learning model compression [39] [40] [41] may need to be performed to minimize the computation workload over the drone.

Fog Computing [42] infrastructure is one of the platform, which could enable deep learning based procedures to be delivered over UAV, which is normally equipped with restricted computation resources. Figure 1 depicts the Fog Computing platform which involves UAV for object detection and environment monitoring. This platform mainly includes three major parts. UAV, together with cameras and sensors are regarded as the sensory infrastructure; Fog/Edge Node could be either laptop computers or mobile phones to undertake intermediate tasks which provide expedite access for end users; the remote cloud center, which possesses the GPU server, could provide training for the deep learning procedures. In such a Fog Computing system, UAV could either execute deep learning inference procedures by itself or send back all the image data to the Fog Node, which provides the further strengthened platform for the deep learning inference procedures to be executed. The cloud center may need to send fine-tuned model to the Fog Node regularly to improve the deep learning prediction precision.

Because of its rich peripheral interfaces, UAV could also be hardware extendable. Previously, custom hardware design such as Field Programmable Gate Array (FPGA) and Digital Signal Processing (DSP) chip have also been used towards UAV to enhance its computation capability[43][44][45]. Furthermore, Application-Specific Integrated Circuit (ASIC) approach could also be explored to implement deep learning inference procedures over UAV.

Besides those fixed computation relocation strategies, the offloading procedures could also be performed in a dynamic way. An adaptive streaming method has been proposed in [6] to utilize UAV based Fog Computing platform for smart surveillances in the city.

### III. EXPERIMENT

The dataset we use for experiment is the Stanford Drone Dataset (SDD) [15]. The SDD contains different categories of objects: pedestrians, bicyclists, skateboarders, cars, buses, and golf carts. We convert the video sequences in SDD to images for CNN training and testing. The algorithm we use is 1) RetinaNet proposed in [38]. The RetinaNet network architecture uses Feature Pyramid Network (FPN) [46] on top of feedforward ResNet architecture [47]. 2) Faster R-CNN, on top of ResNet architecture. 3) SSD, also on top of ResNet architecture. All of the base ResNet models are pre-trained on ImageNet [48].

TABLE I.  EXPERIMENTAL PARAMETERS

| Parameters | Values |
|---|---|
| Initial learning rate | 0.01 |
| Weight decay | 0.0001 |
| Momentum | 0.9 |
| γ value | 2 |
| α value | 0.25 |

Convolutional Neural Network is implemented based on the deep learning framework MXNet [49]. A server with two Intel 10-core Xeon CPU E5-2630 and an NVIDIA Tesla K80 GPU is used for this experiment. The experimental settings are presented in Table I. Stochastic gradient descent (SGD) has been utilized for the training of all deep learning models and horizontal flipping has been employed as a data augmentation strategy. It is to be noticed that γ value and α value are within focal loss function of RetinaNet.

In order to evaluate the effectiveness of leveraging focal loss for object detection, as in [38], Cumulative Distribution Functions (CDF) of the normalized loss are employed to analyze the contribution of foreground and background samples with change of γ values. From Figure 2, it is found that as γ values increase, it exerts minor effect over positive samples; however, it imposes enormous influence over negative samples, which indicates that hard negative samples grant more weight for producing loss values under large γ setting.

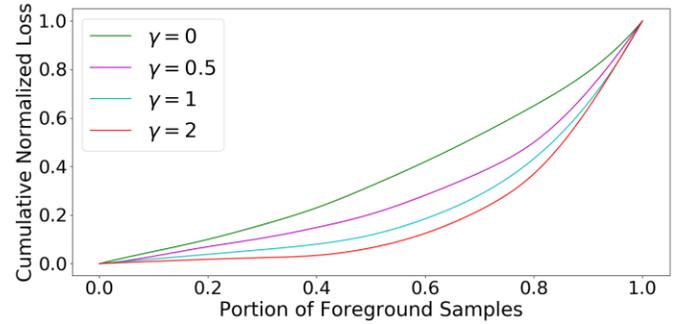

(a) Positive Samples

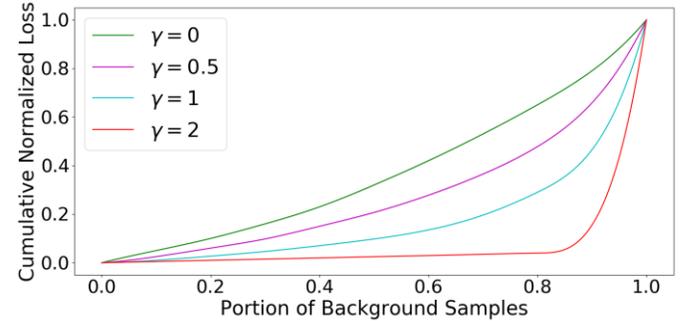

(b) Negative Samples

Figure 2. CDF of the normalized loss for foreground and background samples under different settings of γ values

TABLE II.  ACCURACY AND SPEED RESULT ON SDD.

| method | category | FPS | mAP |
|---|---|---|---|
| SSD(ResNet-50) | one-stage | 23.26 | 80.42 |
| Faster R-CNN (ResNet-50) | two-stage | 13.18 | 83.64 |
| RetinaNet(ResNet-50) | one-stage | **24.45** | 85.17 |
| SSD(ResNet-101) | one-stage | 21.52 | 81.92 |
| Faster R-CNN (ResNet-101) | two-stage | 11.27 | 85.33 |
| RetinaNet(ResNet-101) | one-stage | 22.31 | **86.58** |

Table II showed the final result generated from the comparison of three object detectors. Considering setting up a general baseline, we incorporate ResNet with different depth (50 and 100). We could find that for such SDD dataset based object detection task, RetinaNet could perform faster as well as more accurate in object detection from UAV.

Figure 3 shows the visual effect of using RetinaNet to perform the object detection over the Stanford Drone Dataset. Figure 3 (a) and Figure 3 (b) show different scenes in Stanford Drone Dataset. CNN based method performs well in identifying the objects inside the images.

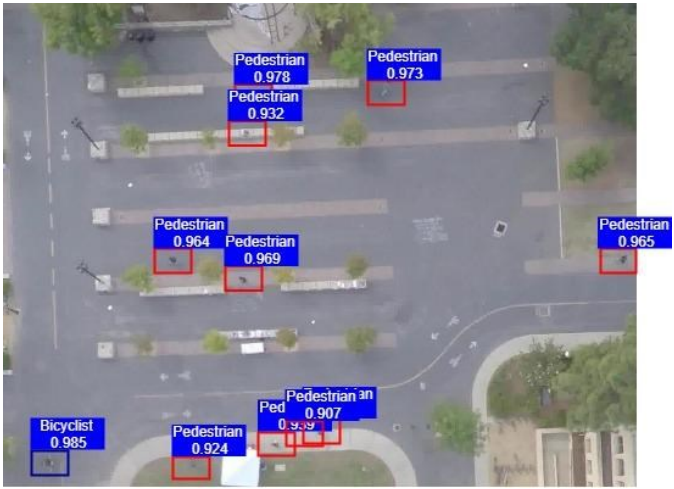

(a)

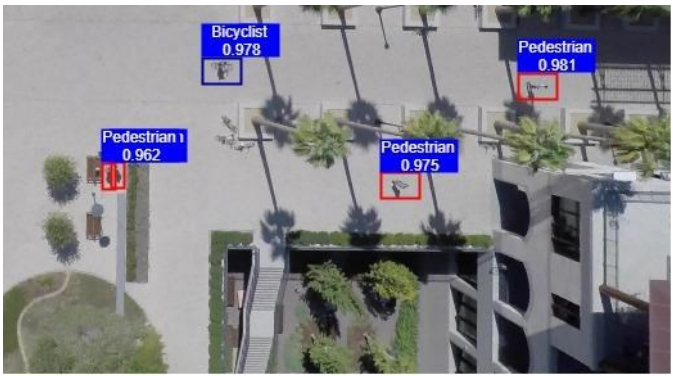

(b)

Figure 3. RetinaNet Object Detection on SDD

## IV. CONCLUSION

We effectively explored the use of Convolutional Neural Network based object detector, especially the lately proposed focal loss dense detector - RetinaNet for object detection from UAV. The experiment result has shown great advantages of leveraging deep learning based methods for executing computer vision tasks from UAV as opposed to other methods. For the next step of the study, we plan to investigate further the use of different implementation platforms to deploy Convolutional Neural Network based methods within UAV system. As illustrated previously, offloading computation from UAV to ground Fog node and Cloud Center or utilizing hardware design to facilitate the deep learning execution on UAV itself would both be worth testifying. With all of these methods, computation intensive deep learning procedures could be migrated to drone based systems to facilitate the real time monitoring of variable complicated situations under different environmental conditions.